# Multi-Unit Association Measures:
# Moving Beyond Pairs of Words

Jonathan Dunn

*Abstract*

This paper formulates and evaluates a series of multi-unit measures of directional association, building on the pairwise *ΔP* measure, that are able to quantify association in sequences of varying length and type of representation. Multi-unit measures face an additional segmentation problem: once the implicit length constraint of pairwise measures is abandoned, association measures must also identify the borders of meaningful sequences. This paper takes a vector-based approach to the segmentation problem by using 18 unique measures to describe different aspects of multi-unit association. An examination of these measures across eight languages shows that they are stable across languages and that each provides a unique rank of associated sequences. Taken together, these measures expand corpus-based approaches to association by generalizing across varying lengths and types of representation.

<u>Keywords</u>: association strength, multi-unit association, sequences, ΔP, collocations

## 1. Introduction

The goal of this paper is to generalize measures of linguistic association across both the direction of association and the number of units in a sequence. Association measures quantify which linguistic sequences co-occur in a significant or meaningful way (e.g., Church & Hanks 1990; Gries & Stefanowitsch 2004; Gries 2008). Traditionally, association has been viewed as a relationship between two lexical items. Given the utterance in (1a), for example, traditional measures represent the degree to which neighboring pairs of words such as (1b) are associated with one another. The problem is that this misses larger phrases such as (1c) that contain more than two words. The point of this paper is to expand the scope of association measures to sequences of varying length and level of representation while maintaining directional distinctions.

> (1a) please give me a hand here
> (1b) give me
> (1c) give me a hand
> (1d) give me a hand here
> (1e) give [NOUN] a hand
> (1f) give her a hand

The expansion to phrases of varying length creates a new problem that pairwise measures implicitly ignore: segmentation. For example, the phrase in (1d) contains the phrase from (1c) as a sub-sequence but also includes *here*. Association measures that are not confined to arbitrary lengths must be able to segment sequences in order to identify phrases like (1c) nested within larger sequences like (1a). Similarly, association can be generalized beyond word-forms to describe sequences such as (1e), in which a partially-filled slot, NOUN, allows greater descriptive generalizations that encompass phrases like (1f) as well as (1c). This creates an additional problem: is (1c) or (1e) the best representation for this phrase?

In order to better understand this problem, the paper develops and evaluates a series of multi-unit directional association measures, each building on the pairwise *ΔP* measure (Ellis 2007; Gries 2013), across eight languages (German, English, Dutch, Swedish, French, Italian, Spanish, Portuguese) at two levels of representation (lexical and syntactic). This evaluation importantly allows us to observe both (i) relationships between the measures and (ii) the stability of their behavior across languages.

## 2. Direction of Association and Sequence Length

Previous approaches to association have been confined to pairs of words like *give me* and *a break* and are thus unable to discover phrases like *give me a break*. In generalizing beyond pairs of words, this paper uses the term 'sequence' to refer to phrases of any length (instead of length-specific terms like 'pair' or 'bi-gram'). In generalizing beyond lexical representations, this paper uses the term 'unit' to refer to words in a sequence, whether represented using their orthographic form (*give*) or using their part-of-speech (VERB).

This paper focuses on five requirements for multi-unit association measures: First, linguistic association is direction-specific. Directional association measures reveal the dominating direction of attraction between units (i.e., Pedersen 1998; Ellis 2007; Michelbacher 2011; Gries 2013). This is important because language is one-dimensional in the sense that *give* in (1a) comes after *please* but before *me*. Thus, association strength is a relationship that can hold in either direction: left-to-right (referred to here as LR) from *please* to *give* and right-to-left (referred to here as RL) from *give* to *please*. Asymmetries in directional association make this an important property that needs to be captured by association measures (c.f., Gries 2013).

Second, traditional association measures deal only with pairs of words (c.f., Evert 2005; Wiechmann 2008; Pecina 2009); meaningful associations, however, may extend to any length (c.f., Jelinek 1990; Daudaravicius & Marcinkeviciene 2004). Further, the division between words in orthography is not principled so that individual words may be treated as multiple units. Thus, another important criteria is that association measures offer comparable quantifications of association regardless of the number of units (i.e., words) in a sequence (i.e., phrase).

Third, while association measures are usually applied to lexical sequences, there are sequences like *give [NOUN] a hand* in which one of the elements in the sequence is partially unfilled. The idea is that a sequence may be associated with any lexical item from a given category (here, any noun; c.f., Gries 2010; Gries & Mukherjee 2010; Wibble & Tsao 2010). Such sequences would appear unassociated given only a single noun. For example, *give him a hand* and *give Joe a hand* are instances of a single idiom but this pattern cannot be captured using purely pairwise lexical association.

Fourth, association measures need to be calculated efficiently. Here this means that only sequence-internal properties can be taken into consideration. For example, the association between *give* and *me* should not be calculated using the association between *give* and anything other than *me*. The problem is that measures which take external information into account (e.g., Shimohata, et al. 1997; Zhai 1997) effectively prohibit the search for associated sequences across many lengths and types of representation.

Fifth, many sequences contain arbitrary segmentations. For example, *give me* from the longer phrase *give me a hand* is an arbitrary segmentation if it follows only from a length restriction. When measuring multi-unit association, we can (i) try to develop measures that are length agnostic so that a single set of measures covers all lengths or (ii) class results by length so that we find the most associated bigrams, trigrams, etc. in independent batches. The goal in this paper is to provide measures that generalize beyond sequence length. This generalization increases the impact of the segmentation problem: given a sequence of associated units, as in (1a) above, how can we determine whether (1a) as a whole is a collocation or whether it contains a sub-sequence like (1c) which is a collocation?

In this paper a sequence is a string of words for which we know only precedence relations. In the sequence *the big red dog*, for example, we observe that *red* comes before *dog*. Such precedence relations are the only ones that we observe (i.e., semantic and syntactic relations are not directly available). An individual 'instance' is one occurrence of a sequence and may occur many times in a large corpus. Association strength, then, is a measure of how meaningful a particular precedence relationship is across instances. Thus, if *red dog* occurs together only a few times relative to the individual frequencies of *red* and *dog*, the precedence relationship that we observe in this particular instance does not generalize to strong association across the corpus. The essential difference between pairwise and multi-unit measures is that pairwise measures require only the concept of co-occurrence (i.e., that *red* and *dog* occur together in an observed string) while multi-unit measures require the additional concept of precedence relations: *the big red dog* is actually a chain of precedence relations that holds across individual pairs. We need to generalize from co-occurrence of units to co-occurrence of precedence relationships between units.

Within a sequence, individual units can be either lexical items (2a) or parts-of-speech (2b). We can generalize across types of representation by referring to sequences of units, not specifying if a particular slot is filled by a lexical item or by (any member of) a syntactic category. An abstract sequence is given in (2c), where each letter indicates a unit (e.g., a lexical item) with dashes separating slots in the sequence (i.e., positions occupied by units).

(2a) the – big – red – dog

(2b) DETERMINER – ADJECTIVE – ADJECTIVE – NOUN

(2c) A – B – C – D – E – F – G

(2d) B – C – D

(2e) C – D – E – F – G

(2f) A – B

(2g) he was about to give me a hand

The advantage of this abstraction is that we can define multi-unit measures without assuming the number of units. First, we need the concept of 'end-points': each sequence has a left and a right end-point: the first and last units in the sequence. For example, the left end-point in (2c) is A and the right end-point is G. Second, we need the concept of 'sub-sequences': any sequence of more than two units can be reduced to one or more contained sequences; for example, the sequence in (2c) includes among others the sub-sequences given in (2d) through (2f). To make the problem of sub-sequences concrete, the sentence in (2g) contains the multi-word idiom *give me a hand* along with a number of sequences like *me a hand* that are not meaningful. The problem for multi-unit measures is to determine where the boundaries of an idiom begin and end. In other words, multi-unit measures must be able to indicate when a sub-sequence is more associated than the sequence as a whole. Third, we need the concept of 'neighboring pair': any two adjacent units within a sequence. Thus, the set of neighboring pairs in (2a) is: *the big*, *big red*, *red dog*.

The core of all the multi-unit measures developed in this paper is the pairwise *ΔP*: Let *X* be a unit of any representation and *Y* be any other unit of any representation, so that $X_A$ indicates that unit *X* is absent and $X_P$ indicates that unit *X* is present. We are concerned with association in both possible directions, left-to-right (LR) and right-to-left (RL). The LR measure is $p(X_P|Y_P)$ - $p(X_P|Y_A)$ and the RL measure is $p(Y_P|X_P)$ - $p(Y_P|X_A)$. This is simply the conditional probability of co-occurrence in the given direction (i.e., of *Y* occurring after *X*) adjusted by the conditional probability without co-occurrence (i.e., of *Y* occurring without *X*). In its original formulation, the $ΔP$ was meant to indicate the probability of an outcome given a cue, $p(X_P|Y_P)$, reduced by the probability of the outcome in the absence of the cue, $p(X_P|Y_A)$. In linguistic terms, the outcome is co-occurrence of two units and the cue is the occurrence of only one of the units. In this paper, the direction of association being measured is notated using a sub-script: left-to right is written as $ΔP_{LR}$ and right-to-left as $ΔP_{RL}$.

For the purposes of illustration, Table 1 defines a schematic co-occurrence matrix that will be used to show how the pairwise $ΔP$ is calculated (for further details, see Gries 2013). This matrix allows an abstraction on top of observed co-occurrences (i.e., strings in which unit $X$ and unit $Y$ occur as $XY$). The number of occurrences of $X$ and $Y$ together is given by $a$. The number of occurrences of X without Y is given by $b$ and of $Y$ without $X$ by $c$. To capture the size of the corpus, the number of units occurring without either $X$ or $Y$ is given by $d$. These four variables allow other quantities to be defined: the total number of occurrences of $X$, for example, is $a + b$ (i.e., its occurrences both with and without $Y$). For the base pairwise measure, the LR conditional probability $p(X_P|Y_P)$ can thus be calculated as $a / (a + c)$ or the number of cases of $X$ and $Y$ occurring together over the total number of cases in which $Y$ occurs. Here, the presence of $Y$ is the conditioning factor and this represents left-to-right association. The RL conditional probability $p(Y_P|X_P)$ can be calculated as $a / (a + b)$ or the number of cases of $X$ and $Y$ occurring together over the total number of cases in which $X$ occurs. Thus, the presence of $X$ is the conditioning factor and this represents right-to-left association. The full formula for $ΔP_{LR}$ is given in (3a) and for $ΔP_{RL}$ in (3b).

*Table 1. Schematic Co-Occurrences for Unit X and Unit Y*

|  | **Y Present ($Y_P$)** | **Y Absent ($Y_A$)** | **TOTALS** |
|---|---|---|---|
| **X Present ($X_P$)** | a | b | a + b |
| **X Absent ($X_A$)** | c | d | c + d |
| **TOTALS** | a + c | b + d |  |

(3a) $ΔP_{LR} = a / (a + c) - b / (b + d)$

(3b) $ΔP_{RL} = a / (a + b) - c / (c + d)$

Consider the phrase *give me a hand*, whose relevant frequencies from the Corpus of Contemporary American English (Davies, 2008) are shown in Table 2. For example, *give* occurs on its own 189,583 times and occurs preceding *me* 15,049 times. These frequencies, when put into a co-occurrence table, give the values shown in Table 3. When put into the formulas in (3), this provides a $ΔP_{LR}$ of 0.015 and a $ΔP_{RL}$ of 0.077. The purpose of this brief example is to show how individual frequencies are abstracted into co-occurrence frequencies which are then used to calculated the $ΔP$ measure in both directions.

*Table 2. Example Word Frequencies*

| Unit | Frequency | Unit | Frequency |
|---|---|---|---|
| give | 189,583 | me a | 28,825 |
| me | 959,874 | a hand | 11,394 |
| a | 12,040,614 | me a hand | 185 |
| hand | 183,627 | give me a | 4,966 |
| give me | 15,049 | give hand | 4 |

*Table 3. Co-Occurrence Frequencies for "give me"*

|  | *me* Present ($Y_P$) | *me* Absent ($Y_A$) | TOTALS |
|---|---|---|---|
| *give* Present ($X_P$) | a = 15,049 | b = 174,534 | a + b = 189,583 |
| *give* Absent ($X_A$) | c = 944,825 | d = 518,865,592 | c + d = 519,810,417 |
| TOTALS | a + c = 959,874 | b + d = 519,040,126 |  |

The main problem addressed in this paper is that current work does not cover multi-unit sequences, does not adequately cover direction-specific measures, does not cover multiple types of representation, and does not adequately examine the behavior of different measures across languages. In order to address these gaps in the literature, the next section introduces a multi-lingual experimental set-up using data from the Europarl Corpus representing eight languages and two levels of representation (lexical and syntactic). The fourth section formulates eight directional multi-unit measures (producing separate LR and RL scores) and two measures that compare directional association (producing only a single score). The fifth section undertakes a large-scale quantitative study in order to answer two basic questions: First, what is the relationship between different directions of the same measure and between different measures in the same direction? Second, how stable are these measures across languages and types of representation? An additional external resource explores the behavior of variants of the measures that (i) use frequency weighting to take sequence frequency into account and (ii) use conditional probability as the base measure rather than the *ΔP*. Finally, the sixth section discusses how these measures can be combined into a single sequence-filtering mechanism.

## 3. Data and Methodology

This paper undertakes a large-scale quantitative study of various multi-unit association measures (introduced in Section 4) in order to answer two basic questions: First, how similar are the ranks of sequences produced by different measures and different directions of association? Do we need all of them? How do they interact? Second, how stable are these measures across different conditions? What factors influence their behavior? This experimental design focuses on the influence of language, type of representation, and sequence length. The external resources also contain an examination of frequency-weighted variants and conditional probability variants of the underlying measures that are beyond the scope of the paper itself.

The first condition is language: does the behavior of each measure vary across languages? We use corpora from eight languages from the Europarl corpus of European Parliament proceedings (Koehn 2005): four Germanic languages (German: *de*, English: *en*, Dutch: *nl*, Swedish: *sv*) and four Romance languages (French: *fr*, Italian: *it*, Spanish: *es*, Portuguese: *pt*), with 650k speeches each. This allows each language to represent the same domain. We consider sequences containing between 2 and 5 units with part-of-speech tagging performed using RDRPosTagger (Nguyen, et al. 2016).

The first question is purely descriptive: how many sequences are there across languages, lengths, and levels of representation? How do frequency and dispersion constraints change the number of sequences? Dispersion, the distribution of a sequence across a corpus (i.e., Gries 2008, Biber, et al. 2016), is implicitly treated by processing the corpus in chunks of 500 speeches; any sequence that falls below a per-chunk frequency threshold (set at 10) is discarded. This favors evenly dispersed sequences while maintaining efficiency. A further individual unit threshold (set at 50) removes sequences that contain infrequent lexical items.

The algorithm for extracting sequences has two passes: first, building an index of individual units and, second, building an index of sequences. In the first pass, all individual words are counted. Infrequent words are discarded; a word must occur roughly once every million words to be indexed. Given the Zipfian distribution of word frequencies, a very large number of less frequent words would need to be indexed without this threshold. The effect of the Zipfian distribution is much greater with multi-unit sequences because there are many more sequences than there are units. The individual frequency threshold means that no sequences containing words below that threshold need to be indexed, reducing the problem of large numbers of infrequent sequences. Because the algorithm processes small batches of the corpus in parallel, each batch also contains a very large number of very infrequent sequences whose total frequency cannot be known until all batches are processed. The per-chunk threshold allows the algorithm to discard infrequent sequences within each batch. The influence of the individual frequency threshold is shown in Figure 1. As this threshold is raised from 500 to 2,000 the number of sequences shrinks quickly, as represented by 'Sequences Before Threshold'. On the other hand, if we enforce an additional sequence frequency threshold of 1,000 the growth is much reduced, as represented by 'Sequences After Threshold'. This means that the individual unit threshold removes a large number of sequences, but that most of these removed sequences are themselves infrequent. In part, this follows from the fact that a given sequence can be no more frequent than its least frequent unit.

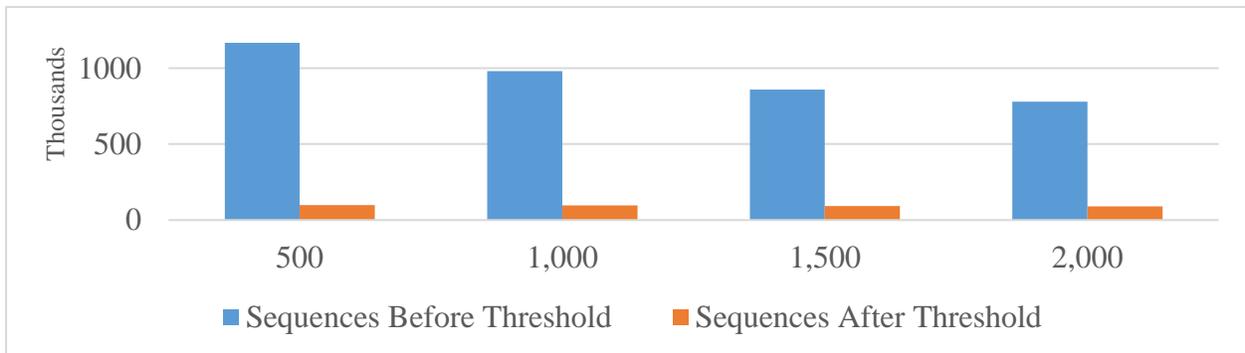

*Figure 1. Influence of Frequency Threshold for Individual Units for English*

Dispersion is implicitly enforced by using a per-chunk sequence frequency threshold to remove those sequences which do not occur in a given part of the corpus; because the corpus is processed in many chunks this reduces the prominence of poorly distributed sequences. The impact of this threshold is shown in Figure 2, where the individual threshold is held constant (at 2,000) and the per-chunk threshold is raised from 2 to 5. As before, two conditions are compared: 'Sequences Before Threshold' is the set of all sequences and 'Sequences After Threshold' is the set of all sequences that occur more than 1,000 times. We see, then, that increasing the per-chunk threshold sharply reduces the total number of sequences in the corpus. However, the per-chunk threshold has a much reduced impact on more frequent sequences. This means, as before, that the thresholds used to maintain efficiency largely impact sequences that occupy the very long tail of the Zipfian distribution. Note that Figures 1 and 2 use different thresholds than all later figures in order to make these comparisons possible.

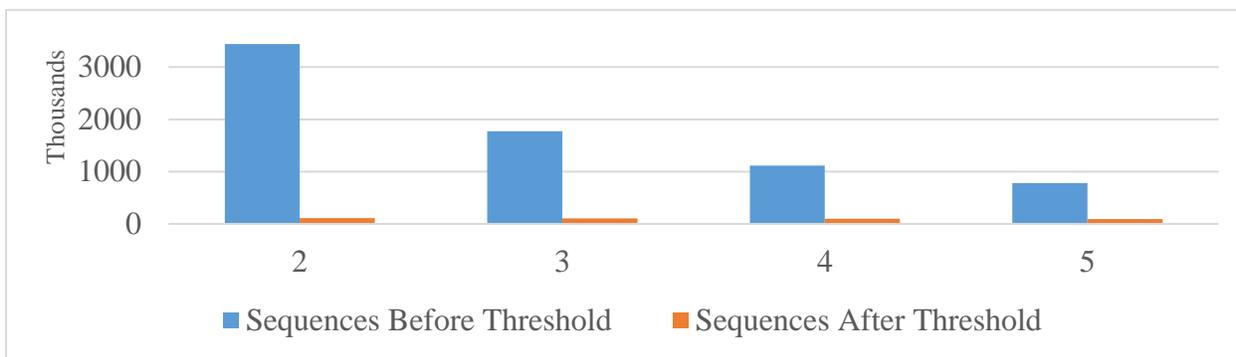

*Figure 2. Influence of Per-Chunk Threshold for Sequences for English*

The number of words contained in 650k speeches is given in Figure 3; because the speeches largely overlap, variations across languages are linguistic in nature. Although relatively similar, the corpora range from 45.9 million words (English) to 54.3 million words (French). The purpose of Figure 3 is to show that we expect some variation in sequences across languages simply as a result of having different numbers of units in the corpus. It turns out, however, that the number of sequences for each language varies more widely than this baseline, from 153k (German) to 298k (French) as shown in Figure 4.

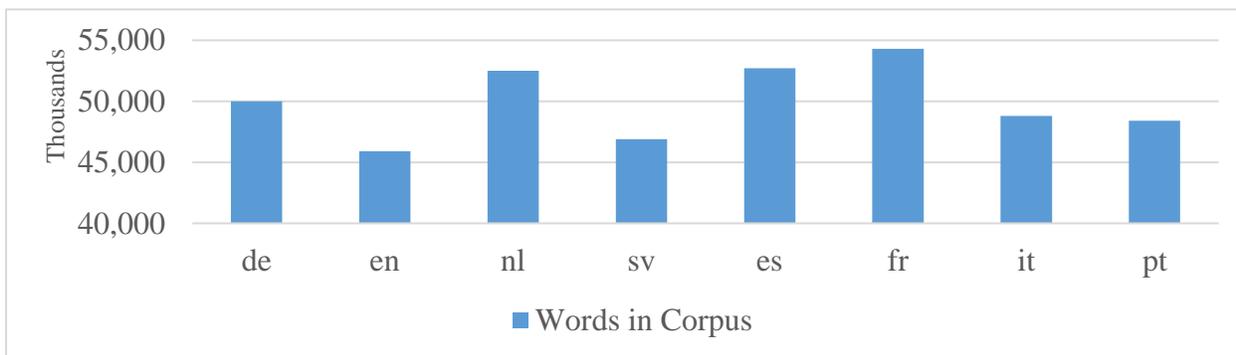

*Figure 3. Number of Words in Corpus Across Languages*

As also shown in Figure 4, the number of sequences when parts-of-speech are included is much higher across all languages than the number of purely lexical sequences. Given that each language has a separate tag set and tagging model, it could be the case that finer-grain tags for some languages produce a larger number of sequences. However, there is also variation in the number of purely lexical sequences: ranging from 14,800 (Swedish) to 29,300 (Spanish). This is visualized in Figure 5 with a closer look at only lexical sequences, showing that tag sets are not the sole cause of this variation. In fact, the distribution of lexical and total sequences largely correspond across languages, again indicating that the number of sequences is more than an artifact of the tag sets. The number of sequences also illustrates the importance of efficiency: the number of sequences grows quickly when length and representation constraints are removed.

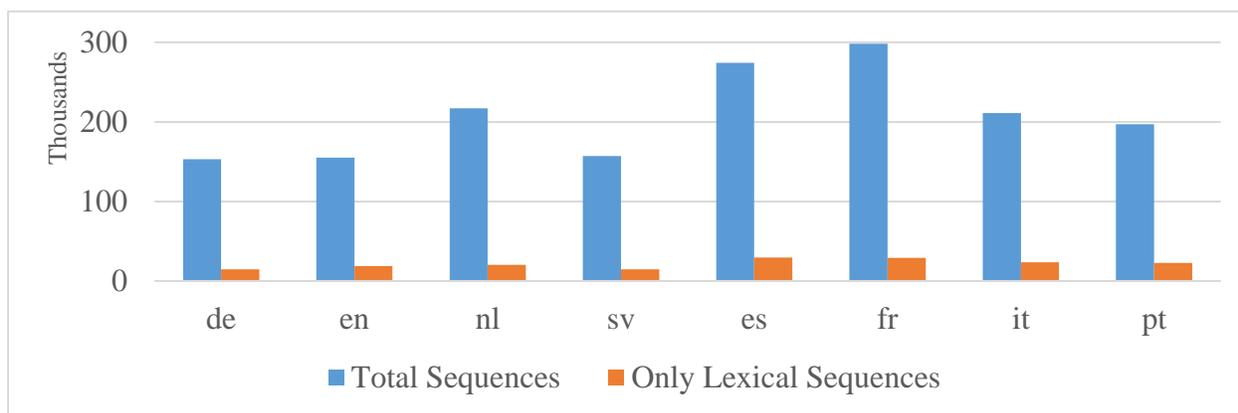

*Figure 4. Number of Sequences in Corpus Across Languages*

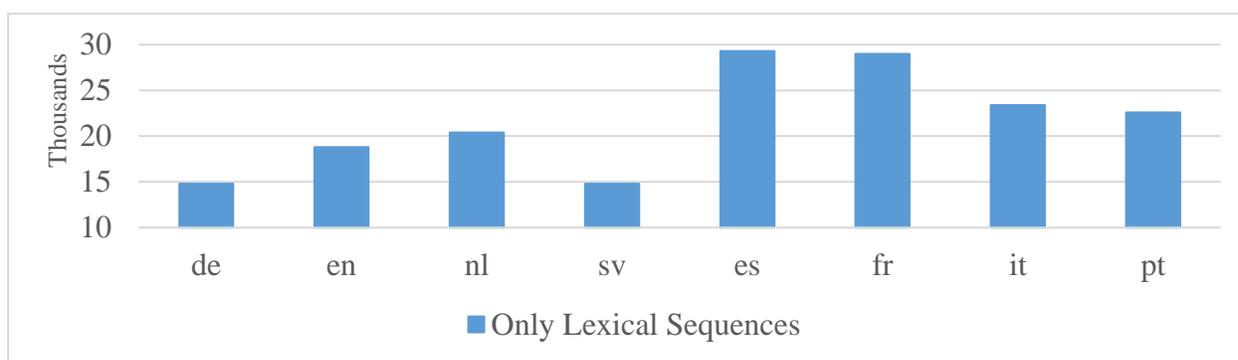

*Figure 5. Number of Lexical Sequences in 650k Speeches Across Languages*

Finally, how does sequence length influence the number of sequences across languages? This is important because the purpose of this paper is to remove the current restriction to pairs of words, a generalization which impacts the number of sequences as shown in Figure 6. Across languages, longer sequences have more types (although the average sequence frequency goes down as length increases). This is the case for each language but the magnitude of the increase varies: the difference between the maximum and minimum number of sequence types across languages is 9k for length 2 but 69k for length 5. This shows, again, that the difference in sequences across languages is greater than the baseline variation in the number of words in each corpus. Further, a high number of sequence types that is caused simply by a higher number of categories (i.e., more word types or parts-of-speech) would result in a lower average sequence frequency. Instead, the opposite is the case, with a higher number of sequence types often co-occurring with a higher average type frequency (not shown). These variations, then, reflect differences across languages that justify the empirical examination of association measures across languages even though including language as a dimension of variation complicates the analysis. A list of the parameters used for the experiments described in the following sections is given in Table 4.

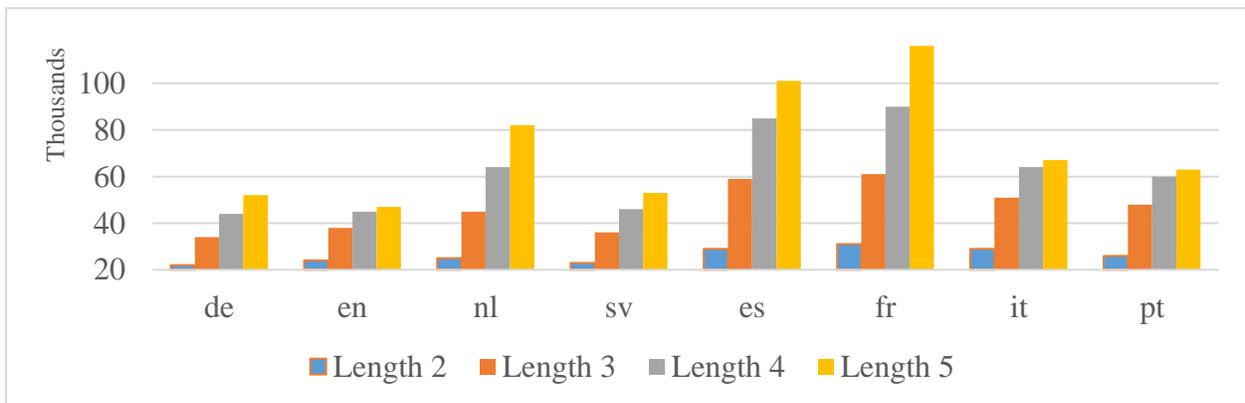

*Figure 6. Number of Sequences in 650k Speeches Across Languages By Length*

*Table 4. Parameters*

| Name | Value | Description |
| --- | --- | --- |
| Individual Frequency | 50 | Threshold for individual units across entire corpus |
| Per-Chunk Frequency | 10 | Threshold for sequences within individual chunks |
| Chunk Size | 500 | Number of speeches included in each chunk of corpus |

## 4. Analysis: Formulating Multi-Unit Association Measures

### 4.1. Mean ΔP and Sum ΔP

A multi-unit sequence can be viewed as a sequence of neighboring pairs. Our first two measures represent the accumulation of pairwise association across units in a sequence: the Mean *ΔP*, notated as *μ(ΔP$_{LR}$)*, and the Sum *ΔP*, *Σ(ΔP$_{LR}$)*. Both measures are calculated across neighboring pairs, formalized in (4b) and (4c), where *{NP}* indicates the set of neighboring pairs and *n{NP}* indicates the number of neighboring pairs. For example, given the sequence in (4d), these measures simply take the mean and the sum of each pairwise *ΔP*: *european union*, *union solidarity*, and *solidarity fund*. When calculated from the Europarl corpus, the sequence in (4d) has an *μ(ΔP$_{LR}$)* of 0.647 and an *μ(ΔP$_{RL}$)* of 0.277; thus, it exhibits the sort of asymmetric association that makes direction-specific measures important. The sequence in (4e), on the other hand, has values of -0.062 and -0.028: it is equally unassociated in both directions.

(4a) *NP* = Set of Neighboring-Pairs in Sequence
(4b) *Σ(ΔP$_{LR}$)* = Σ *ΔP$_{LR}${NP}*
(4c) *μ(ΔP$_{LR}$)* = Σ *ΔP$_{LR}${NP}/n{NP}*
(4d) european – union – solidarity – fund
(4e) of – living – and – working

These are simple methods for combining pairwise association values. *Σ(ΔP)* favors longer sequences because association accumulates as more units are added; thus, *Σ(ΔP)* works best in a scenario in which sequences are compared by length (i.e., trigrams against trigrams). For example, Table 4 shows the top lexical sequences for the *Σ(ΔP)* measure in the LR direction: most sequences contain several units. On the other hand, the *μ(ΔP)* tends to favor shorter sequences because it looks at the average association across neighboring pairs: highly associated pairs will rise to the top and longer sequences will have difficulty matching simple pairs. In both cases, length's main influence is that longer sequences can better tolerate weak links. For example, the top sequences for *Σ(ΔP$_{LR}$)* include two phrases that contain *in order to make* and six that contain *of the european union*. The high association of these sub-sequences tends to promote any sequence that contains them. This illustrates the segmentation problem: when is a sub-sequence better than the sequence as a whole? Note that capitalization is not used in Table 5; this is because association measures are calculated on lower-case representations.

Table 5. Top 10 Lexical Sequences for $\mu(\Delta P)$ and $\Sigma(\Delta P)$ in the LR Direction

| $\mu(\Delta P_{LR})$ | | $\Sigma(\Delta P_{LR})$ | |
|---|---|---|---|
| and gentlemen | 0.888 | i am voting in favour | 1.454 |
| the same | 0.797 | in order to make it | 1.245 |
| in order | 0.758 | in order to make | 1.245 |
| i am | 0.741 | i would like to say | 1.151 |
| in favor | 0.686 | implementation of the european union | 1.081 |
| liu xiaobo | 0.686 | of the european union and | 1.080 |
| of course | 0.658 | mobilisation of the european union | 1.079 |
| member states | 0.605 | functioning of the european union | 1.079 |
| european union | 0.599 | of the european union | 1.079 |
| porto alegre | 0.585 | membership of the european union | 1.079 |

What is the relationship between the sequence rankings produced by the $\mu(\Delta P)$ and $\Sigma(\Delta P)$ measures? The top sequences in Table 5 suggest that the two measures produce significantly different rankings, but this is actually not the case. Quantitative relationships between these and other measures, however, will be considered in more detail only in Section 5.

*4.2. Minimum $\Delta P$*

The first two multi-unit measures reveal the problem of sub-sequences: many of the top sequences contain the same neighboring pairs. The Minimum $\Delta P$, or $M(\Delta P)$, tries to identify weak links within a sequence. The idea is that such weak links provide a quick check to see if a sequence contains unassociated material. For example, if the $\Sigma(\Delta P)$ of *in order to make* and *in order to make it* are the same (at this level of precision), then the final pair *make it* clearly does not add to the overall association of this sequence: it is a weak link. This is formalized in (5), where *NP* is again the set of neighboring pairs in a sequence and $M(\Delta P_{LR})$ is simply the minimum observed $\Delta P_{LR}$ across all neighboring pairs. Thus, the $M(\Delta P)$ on its own is not an association measure because it simply finds the weakest link in a chain of pairwise association values. But, when combined with other measures, it provides a way of filtering out sequences with segmentation problems.

(5a) *NP* = Set of Neighboring-Pairs in Sequence

(5b) $M(\Delta P_{LR}) = \min(\Delta P_{LR}\{NP\})$

For example, Table 6 shows the same rankings as Table 5, this time with sequences containing a $M(\Delta P_{LR})$ of less than 0.01 removed. The $\mu(\Delta P_{LR})$ in Table 5 had only individual pairs among its top sequences; it turns out that $M(\Delta P)$ is always the same as $\mu(\Delta P_{LR})$ and $\Sigma(\Delta P_{LR})$ for sequences of length two. Thus, only the filtered and unfiltered top ranked sequences for $\Sigma(\Delta P_{LR})$ are shown in Table 6. On the one hand, the repeated sub-sequences from Table 5, many of which contained weak links, have been filtered out. On the other hand, those phrases which have risen to the top still contain core sub-sequences (for example: *of, in, on, for, that* with *the european union*). This is important because the goal is to remove sub-sequences that result from poor segmentation, not sub-sequences that form their own collocations.

*Table 6. Top 10 Lexical Sequences for $\Sigma(\Delta P)$ in the LR Direction, Filtered By $M(\Delta P)$*

| *Filtered* | *M($\Delta P$)* | *Unfiltered* | *M($\Delta P$)* |
|---|---|---|---|
| in order to make | 0.017 | i am voting in favour | 0.000 |
| i would like to say | 0.039 | in order to make it | 0.000 |
| of the european union | 0.053 | in order to make | 0.017 |
| i would like to thank | 0.039 | i would like to say | 0.039 |
| in the european union | 0.038 | implementation of the european union | 0.002 |
| on the european union | 0.053 | of the european union and | 0.000 |
| in order to achieve | 0.017 | mobilisation of the european union | 0.000 |
| for the european union | 0.019 | functioning of the european union | 0.000 |
| that the european union | 0.016 | of the european union | 0.053 |
| of the european arrest warrant | 0.053 | membership of the european union | 0.000 |

*4.3. Reduced $\Delta P$*

Even with weak links removed by filtering with $M(\Delta P)$, we still face the problem of independently associated sub-sequences. Consider the sequences in (6), all of which have a high $\Sigma(\Delta P_{LR})$ and all of which pass the $M(\Delta P)$ filtering constraint. With access to introspection, we can see this as a large collocation in (6c/d) that contains a smaller collocation in (6a) as well as a smaller but incomplete sequence in (6b). How can we distinguish between independently associated sequences like (6a) and incomplete sub-sequences like (6b)? The problem is that all of the sequences in (6) are ranked highly by the measures we have so far. The Reduced $\Delta P$ class of measures allows us to make a distinction between these sequences by comparing the $\Sigma(\Delta P)$ of a sequence (e.g., *i would like to say*) with its immediate sub-sequence (e.g., *i would like to*).

(6a) i would like
(6b) i would like to
(6c) i would like to say
(6d) i would like to thank

The Reduced class of measures thus works by comparing the $\Sigma(\Delta P)$ of a sequence with and without one of its end-points. This is formalized in (7), where $EP_L$ for *i would like* is *i* and for $EP_R$ is *like*. There are two variants, the Beginning-Reduced and End-Reduced measures that focus on $EP_L$ and $EP_R$, respectively, and are notated as $R_B(\Delta P_{LR})$ and $R_E(\Delta P_{LR})$. As given in (7b), this means that the $\Sigma(\Delta P)$ of *i would like to say* is reduced by the $\Sigma(\Delta P)$ of *i would like to*. How much additional association does the longer sequence provide? Here we are looking at the End-Reduced variant (i.e., 7b and 7d) because the sequences in (6) share their left end-points (i.e., *i would*). Given the sequence in (7c), the formula is shown for $R_E(\Delta P)$ in (7d) and for $R_B(\Delta P)$ in (7e). The difference between these variants is in the sub-sequences they are comparing.

(7a) $EP_L$ = Left end-point of sequence *S*
(7b) $R_E(\Delta P_{LR}) = \Sigma \Delta P_{LR}\{S\} - \Sigma \Delta P_{LR}\{S \text{ without } EP_R\}$
(7c) A – B – C – D – E
(7d) $R_E(\Delta P_{LR}) = \Sigma \Delta P_{LR}\{ABCDE\} - \Sigma \Delta P_{LR}\{ABCD\}$
(7e) $R_B(\Delta P_{LR}) = \Sigma \Delta P_{LR}\{ABCDE\} - \Sigma \Delta P_{LR}\{BCDE\}$

If a sub-sequence has a higher mean association value than the full phrase, this measure will have a value near or even below zero. The closer the value is to zero, the more the full sequence represents a poor segmentation. For example, the phrase in (6a) has an $R_E(\Delta P_{LR})$ of 0.414, showing that it improves upon its immediate sub-sequence. The phrase in (6c) has an $R_E(\Delta P_{LR})$ of 0.380, showing that it also improves upon its immediate sub-sequence. However, the incomplete phrase in (6b) has a much lower $R_E(\Delta P_{LR})$ of only 0.039. We see from this example a case of multiple nested sequences, none of which have obvious weak links but which we still need to distinguish between. The Reduced class of measures allows us to quantify this aspect of association, giving a high ranking to (6a) and (6c) but a low ranking to (6b).

The top lexical sequences for $\Sigma(\Delta P_{LR})$ are shown in Table 7, again filtered using $M(\Delta P)$ with a threshold of 0.01 to remove obvious weak links; the table is sorted by $\Sigma(\Delta P_{LR})$ but shows both $R_E(\Delta P_{LR})$ and $R_B(\Delta P_{LR})$ in order to support a comparison of these measures. First, the Reduced measures do not follow directly from the $\Sigma(\Delta P)$: for example, *i would like to say* and *of the european union* have a similar $\Sigma(\Delta P)$ but have a quite different $R_B(\Delta P)$. In this case, the low $R_B$ value for *of the european union* follows from the fact that the core phrase *the european union* is more highly associated than the entire sequence. In fact, sequences with some unit added onto the core phrase *the european union* all have a low $R_B$. On the other hand, sequences whose core starts at the left end-point and contain variable units on the right end-point, such as *in order to achieve*, have a lower $R_E$. In general, the higher a sequence's value for the Reduced class of measures the better its segmentations are.

*Table 7. Top 10 Lexical Sequences for $\Sigma(\Delta P_{LR})$ Filtered by $M(\Delta P)$*

| Sequence | $\Sigma(\Delta P_{LR})$ | $R_B(\Delta P_{LR})$ | $R_E(\Delta P_{RL})$ |
|---|---|---|---|
| in order to make | 1.245 | 0.758 | 0.468 |
| i would like to say | 1.151 | 0.316 | 0.380 |
| of the european union | 1.079 | 0.053 | 0.599 |
| i would like to thank | 1.065 | 0.316 | 0.294 |
| in the european union | 1.064 | 0.038 | 0.599 |
| on the european union | 1.054 | 0.028 | 0.599 |
| in order to achieve | 1.050 | 0.758 | 0.274 |
| for the european union | 1.045 | 0.019 | 0.599 |
| that the european union | 1.042 | 0.016 | 0.599 |
| the european union | 1.025 | 0.426 | 0.599 |

*4.4. Divided $\Delta P$*

So far we have viewed multi-unit association as a sequence of pairwise association values. What if we instead view it as a chain of precedence relations? The pairwise approach has no memory: so long as each link is strong, the likelihood of a particular series of links (i.e., the phrase in 8a) is not taken into account. How can we find sequences whose association as a chain of precedence relations is stronger than its purely pairwise association? The Divided class of measures is our first attempt to capture chains of precedence relations by viewing each sequence as a pair consisting of one end-point and the rest of the sequence as a single unit. For example, the phrase in (8a) has a relatively low $\Sigma(\Delta P_{LR})$ of 0.241. It also has low values for the Reduced class of measures because, when viewed as a series of independent pairs, none of its links are particularly strong. However, if we view it as a sequence of precedence relations, the sequence as a whole is associated. The Divided class of measures quantifies the attraction between *the* and *former yugoslav republic* (notated as $D_B(\Delta P)$) and the attraction between *the former yugoslav* and *republic* (notated as $D_E(\Delta P)$).

(8a) the former yugoslav republic
(8b) Sequence: A – B – C – D – E
(8c) $D_B(\Delta P) = \Delta P\{A|BCDE\}$
(8d) $D_E(\Delta P) = \Delta P\{ABCD|E\}$

The Divided class is defined formally in (8): given the sequence of units in (8b), the $D_B$ measure makes a pair out of the left end-point and the remainder of the sequence: (A|BCDE). The $D_E$ measure makes a pair out of the right end-point and the remainder of the sequence: (ABCD|E). These represent the conditional probability of encountering the remainder of the sequence when given part of the sequence. The idea is that strong collocations can be quantified by how much one end-point selects the remainder of the sequence. Going back to the phrase in (8a), the individual pairwise links between these units are weak, as discussed above; *the former* has a pairwise association (LR) of 0.011. However, given *former yugoslav republic* it is very likely to have been preceded by *the*; and given *the former yugoslav* it is very likely to be followed by *republic*. As a result, this phrase is highly ranked by $D_B$ but not by the measures previously discussed.

The top ranked sequences from $D_B(\Delta P_{LR})$ are shown in Table 8, again using $M(\Delta P)$ to filter out weak links. Each of these phrases has a very high value for $D_B$ but would not have been captured as a collocation given only a series of pairwise links (i.e., their $\Sigma(\Delta P)$ is rather low). Further, neither of the Reduced measures is able to capture these non-pairwise patterns, as shown by the generally low values for $R_B(\Delta P)$ and $R_E(\Delta P)$ in Table 8.

*Table 8. Top Lexical Sequences for $D_B(\Delta P_{LR})$ and $D_E(\Delta P_{LR})$, Filtered By $M(\Delta P)$*

| Sequence | Σ(ΔP) | $R_B(\Delta P)$ | $R_E(\Delta P)$ | $D_B(\Delta P)$ |
|---|---|---|---|---|
| illegally staying third-country | 0.073 | 0.061 | 0.012 | 0.999 |
| genetically modified soya | 0.328 | 0.275 | 0.052 | 0.999 |
| genetically modified foods | 0.287 | 0.275 | 0.012 | 0.999 |
| passenger name record | 0.022 | 0.010 | 0.012 | 0.999 |
| data protection supervisor | 0.153 | 0.021 | 0.131 | 0.999 |
| internal market scoreboard | 0.128 | 0.096 | 0.032 | 0.999 |
| general motors belgium | 0.453 | 0.442 | 0.010 | 0.999 |
| agree with the rapporteur | 0.465 | 0.010 | 0.437 | 0.999 |
| human rights violations | 0.313 | 0.299 | 0.013 | 0.999 |
| committee on budgets | 0.205 | 0.049 | 0.156 | 0.998 |

### *4.5. End-Point ΔP*

Viewing a sequence as a series of precedence relationships, rather than as a series of pairwise associations, allows us to capture additional meaningful multi-unit sequences with the Divided class of measures. But these measures continue to overlook sequences in which the end-points themselves are highly associated but allow variable sequence-internal patterns. For example, the phrase in (9a) has relatively low values for all previous measures. However, it is an instance of a more general template in (9b) which has many examples in this dataset: *the security council*, *the human rights council*, *the transatlantic economic council*. These other instances of this template are ranked highly by one or more other measures, but the less common instance in (9a) is not. At the same time, because it belongs to a common template, we want a measure capable of capturing the fact that these particular end-points (*the* and *council*) accept varying internal units.

(9a) the governing council
(9b) the – [ADJECTIVES] – council

The End-Point class of measures, notated as *E(ΔP)*, uses the pairwise association between the end-points to measure this: ($EP_L$ | $EP_R$) or, in this case, (*the* | *council*). The idea is that a class of interesting sequences contains specific end-points but has dynamic or flexible members internally. For the sequence in (10a), the end-points are defined as in (10b) and (10c), so that the pairwise association between the end-points in sequences of varying length can be defined as in (10d). In this case, if the end-points are not observed to co-occur, a value of zero is given.

(10a) Sequence: A – B – C – D – E

(10b) $EP_L$ = Left End-Point (here, A)

(10c) $EP_R$ = Right End-Point (here, E)

(10d) $E(\Delta P) = \Delta P\{(EP_L \mid EP_R)\}$

Given this constraint, that the end-points must in fact co-occur, the $E(\Delta P)$ has limited coverage: in the Europarl dataset for English, only 650 multi-unit lexical sequences have co-occurring end-points (out of 8,850 lexical sequences with more than two units). Of these, however, 611 would not have been ranked highly on the previous tables, either because none of the measures ranks them highly or because they contain a weak pairwise link. Filtering for weak links is unnecessary here because sequence-internal association is irrelevant. Thus, none of the select sequences in Table 9 would have been identified as a meaningful sequence without the $E(\Delta P)$. Note that all sequences which share a pair of end-points receive the same value for this measure (e.g., *the second world* and *the arab world* both receive a value of 0.353).

Table 9. Select Lexical Sequences for $E(\Delta P_{LR})$, Unfiltered

| | |
|---|---|
| the request for the commission | 0.572 |
| the president of the commission | 0.572 |
| the proposal for a council | 0.478 |
| the transatlantic economic council | 0.478 |
| the temporary committee | 0.443 |
| the economic and social committee | 0.443 |
| the second world | 0.353 |
| the arab world | 0.353 |
| the reform treaty | 0.222 |
| the draft treaty | 0.222 |

The sequences shown in Table 9 are selected, rather than showing the full ranking, because there are many variations on these templates: for example, the full ranking includes 18 types of councils. What we see, however, is that a significant number of meaningful sequences with varying internal structure would have been left unidentified because of weak links between their internal units. At the same time, the $E(\Delta P)$, without the check on weak links, does promote a few poor segmentations. For example, *the house of european* and *the new european* are relatively highly ranked (0.426) even though they would seem to be chopping off an important part of the sequence. Regardless, this measure allows us to capture another facet of multi-unit association.

*4.6. Changed ΔP*

When we view a sequence as a series of pairs, each of which has two directions of association, we can measure the dominating direction of association for a sequence. The final two measures represent stability in the dominating direction of association. There are two variants of this measure: the Changed-Scalar *ΔP*, or $C_S(\Delta P)$, provides the cumulative difference in directional association across all neighboring pairs; the Changed-Categorical variant, or $C_C(\Delta P)$, counts the number of times the dominating direction of association changes across neighboring pairs in the sequence. For example, in (11a) both measures operate over the set of neighboring pairs: *the european*, *european globalisation*, *globalisation adjustment*, and *adjustment fund*. For each pair, the difference between directional associations is given by (11c), where $P_D$ represents the directional difference for a given pair. $C_S(\Delta P)$ can then be defined formally as in (11d), where it is the sum of pairwise directional differences. Given that LR measures are positive, high values indicate a large dominance of left-to-right association; values near zero represent cases of inconsistent directional association; large negative values indicate a large dominance of right-to-left association.

(11a) the – european – globalisation – adjustment – fund
(11b) *NP* = Set of Neighboring-Pairs in Sequence
(11c) $P_D = \Delta P_{LR} - \Delta P_{RL}$
(11d) $C_S(\Delta P) = \sum P_{D\,\{NP\}}$

The point of this measure is to identify sequences with an interaction between directions of association. One weakness of the formulation in (11d) is that pairs containing a highly dominant LR value will cancel out pairs containing a highly dominant RL value; the $C_S$ will then come out close to zero. Thus, the $C_C$ measure counts the number of times the dominating direction changes: let $L_D = 1$ if $P_D > 0$ and $R_D = 1$ if $P_D < 0$. Thus, $C_C = min(L_D, R_D)$. For example, Table 10 shows each neighboring pair from (11a) with its LR and RL association value and their $P_D$. The $C_S$ simply sums the $P_D$ column, for a value of 0.447. The $C_C$ is the number of occurrences of the least common direction. Here, only one pair has a negative value (indicating a dominant RL association), so that the direction changes only once (i.e., $C_C = 1$).

*Table 10. Calculating $C_S$ and $C_C$*

| Neighboring Pairs | $\Delta P_{LR}$ | $\Delta P_{RL}$ | $P_D$ |
|---|---|---|---|
| the european | 0.426 | 0.035 | 0.391 |
| european globalisation | 0.112 | 0.001 | 0.111 |
| globalisation adjustment | 0.310 | 0.134 | 0.176 |
| adjustment fund | 0.044 | 0.275 | -0.231 |

The first use for these measures is as an additional filtering mechanism. For example, Table 11 shows the highest ranked sequences for $\Sigma(\Delta P_{LR})$ filtered by both $M(\Delta P_{LR})$ with a threshold of 0.01 and by $C_C(\Delta P)$, including only 0 values: this gives us sequences with a high pairwise association without weak links that do not have alternating directions of dominance. Comparing this with the top ranked sequences filtered only for weak links, we see that peripheral units are removed. For example, *the european union* is a highly associated one-directional phrase; without filtering by $C_C$, however, this phrase occurs with a number of additional left endpoints: *of, in, on, for, that*. Thus, we can use this measure to distill longer phrases into core sequences by limiting the number of changes in dominating direction of association.

*Table 11. Top Lexical Sequences for $\Sigma(\Delta P_{LR})$*

| Filtered by $C_C$ and $M$ | Filtered only by $M$ |
|---|---|
| the european union | in order to make |
| ladies and gentlemen | i would like to say |
| the european arrest warrant | of the european union |
| the european globalisation adjustment fund | i would like to thank |
| the same time | in the european union |
| the european globalisation adjustment | on the european union |
| at the same | in order to achieve |
| the member states | for the european union |
| and i am | that the european union |
| in order to | of the european arrest warrant |

As discussed above, large positive values for the $C_S$ measure indicate dominant LR association and large negative values indicate dominant RL association. Another use for this measure, then, is to place both directions of association onto a single scale: the top of the scale represents high LR association and the bottom of the scale represents high RL association. Table 12 shows the top and bottom of the rankings produced by $C_S$ (the top sequences are filtered by $M(\Delta P_{LR})$ and the bottom sequences by $M(\Delta P_{RL})$). Although asymmetries in directional pairwise association was one of the starting points for this paper, this is the first time we have been able to use these asymmetries to our advantage in distinguishing between different directions of association on a single scale.

*Table 12. Top and Bottom Lexical Sequences for $C_S(\Delta P)$, Filtered by $M(\Delta P)$*

| Top (LR) | Bottom (RL) |
|---|---|
| the same time | ladies and gentlemen |
| in order to | behalf of the |
| the european globalisation adjustment fund | order to ensure |
| i am voting | order to make |
| to ensure that | like to say |
| at the same time | favour of the |
| the european arrest warrant | favour of this |
| to be able to | favour of this report |
| the commission has | nace revision 2 |
| that the european arrest warrant | believe that it |

*4.7. Summarizing the Association Measures*

Table 13 summarizes these main classes of multi-unit measures and encapsulates the patterns which each is able to discover. The basic idea is that different aspects of multi-unit association can be captured by a number of different measures. First, some sequences contain continuously associated neighboring pairs. But, second, this over-identifies sequences that contain one weak link disguised by other very strong links. Third, some sequences contain independent sub-sequences that reveal poor segmentations. Fourth, some sequences have associated chains of precedence relations that are not necessarily associated as individual pairs. Fifth, some sequences contain varying internal material that reduces neighboring pairwise association. Sixth, some sequences change directions of association, so that a single direction-specific scale does not capture their overall association.

*Table 13. Summary of Measures*

| | | |
|---|---|---|
| $\Sigma$ | LR, RL | Sequences with consistent neighboring pairwise association |
| $\mu$ | LR, RL | Sequences with consistent neighboring pairwise association |
| $M$ | LR, RL | Non-sequences revealed by weak links |
| $R_B, R_E$ | LR, RL | Sequences containing independent sub-sequences |
| $D_B, D_E$ | LR, RL | Sequences with associated chains of precedence relations |
| $E$ | LR, RL | Sequences with fixed end-points but varying internal units |
| $C_S, C_C$ | | Sequences with a single dominating direction of association |

We now calculate all of these measures for the phrase *the european union budget*, which consists of three neighboring pairs shown in Table 14 with their LR and RL values. Also shown in Table 14 are the end-points (*the budget*) and two sub-sequences (*the european union* and *european union budget*); for these multi-unit sequences the $\Sigma(\Delta P)$ is given. Starting with $\Sigma(\Delta P)$, we simply sum the values for all neighboring pairs (i.e., the first three rows of the table) for each direction: LR = 1.025 and RL = 0.263. The $\mu(\Delta P)$ is normalized by the number of pairs (three): LR = 0.341 and RL = 0.087. The final link for this phrase is very weak (with 0.000 association at this precision), and this weak link lowers the $\mu(\Delta P)$ drastically. The $M(\Delta P)$, as expected, finds weak links: LR and RL are both 0.000 because *union budget* has this same low value in both directions.

*Table 14. Relevant Pairs in "the european union budget"*

| Phrase | $\Delta P_{LR}$ | $\Delta P_{RL}$ | $P_D$ |
|---|---|---|---|
| the european | 0.426 | 0.035 | 0.391 |
| european union | 0.599 | 0.228 | 0.371 |
| union budget | 0.000 | 0.000 | 0.000 |
| the budget | 0.054 | 0.000 | -- |
| the european union | 1.025 | 0.264 | -- |
| european union budget | 0.599 | 0.228 | -- |

For the Reduced class of measures, we need the $\Sigma(\Delta P)$ of different sub-sequences. For the $R_B$ variant, this is given by *the european union budget* minus *the european union*: because the final pair *union budget* adds no association, LR = 1.025 – 1.025 and RL = 0.263 – 0.263. In both cases, the zero value indicates a poor segmentation. For the $R_E$ variant, this is given by *the european union budget* minus *european union budget*: LR = 1.025 – 0.599 and RL = 0.264 – 0.228. This variant tests the initial segmentation (*the european*), which holds up well in the LR direction with a value of 0.426.

For the Divided class of measures, we need the *ΔP* of *the* and *european union budget* for the $D_B$ variant, where we take *european union budget* as a single unit. This is not calculated directly from neighboring pairs, but has the following values: LR = 0.920 and RL = 0.000. For the $D_E$ variant, we need the *ΔP* of *the european union* as a single unit and *budget*: LR = 0.000 and RL = 0.000. These are the only measures which require information not shown in Table 14.

The End-Point measures take the association between *the* and *budget* regardless of whatever units fall between them: LR = 0.054 and RL = 0.000. The Changed measures are calculated using the values for each neighboring pair in the $P_D$ column (which is simply LR minus RL). $C_S(\Delta P)$ sums these for a value of 0.762, showing that LR association dominates for this sequence. $C_C(\Delta P)$ finds the number of times that the dominating direction changes across neighboring pairs, for a value of 0. The point of this example has been to show how all the measures are calculated for a single sequence.

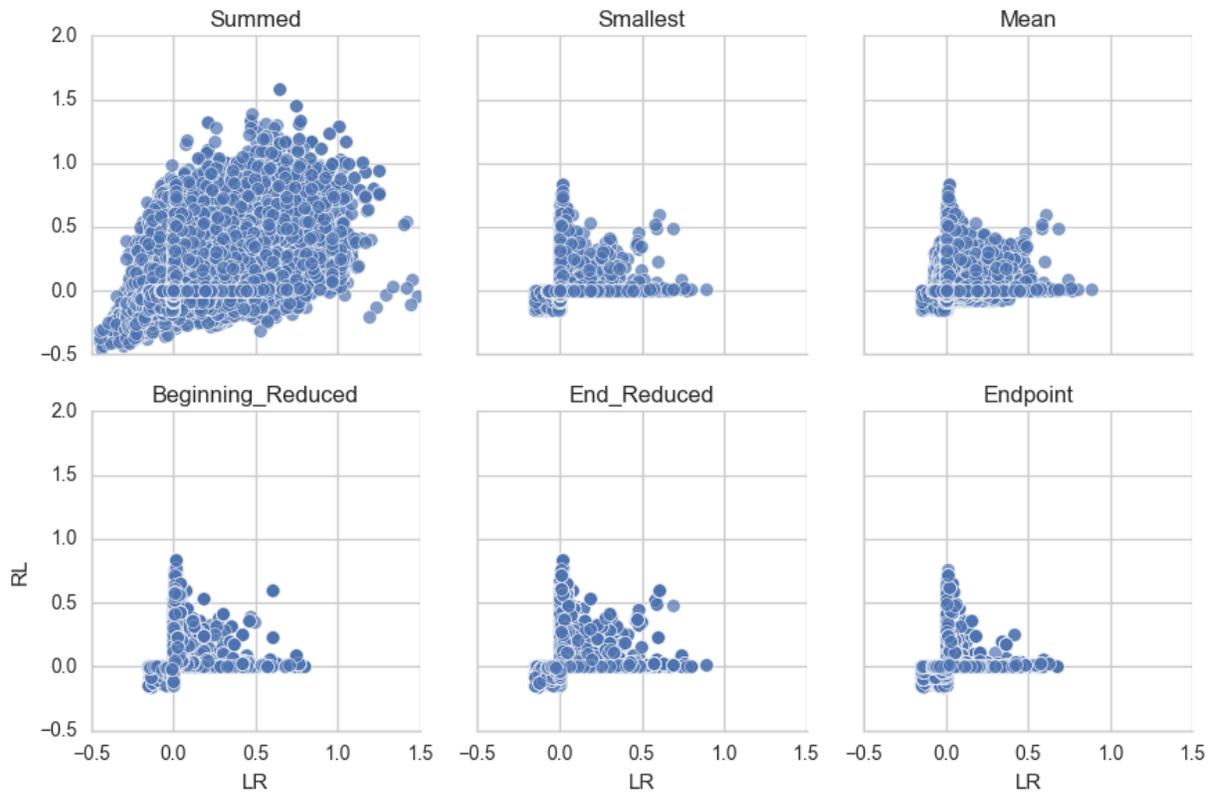

*Figure 7. Scatterplots of LR and RL Point Relations for English*

Do we really need this many measures and directions to capture different aspects of multi-unit association? We can divide this into two simpler questions: First, do the left-to-right and right-to-left measures actually produce different sequence rankings in this larger experiment in the same way they have in smaller previous experiments (i.e., Gries, 2013)? Second, do each of the measures, assuming a single direction of association, actually produce sequence rankings different enough to justify using all of them together, thus forcing the complications of a vector-based approach?

Figure 7 shows scatter plots for each direction-specific measure (excluding the Divided class of measures which are treated separately below). The x-axis plots left-to-right association and the y-axis plots right-to-left association. The $\Sigma(\Delta P)$ has a very slight linear relationship, but the others do not. Note that these plots show only sequences with more than two units because some measures like $R_B(\Delta P)$ have null values for sequences of two units. The corner-like shape produced by most of the measures shows that values near zero for one direction (the most common class of values, since most sequences are not associated) do not predict a similar value in the other direction. This means that, while the distributions across directions is the same (the figure showing this is provided for reference in the external resources), for individual sequences there is not a linear relationship between left-to-right and right-to-left association.

We now turn to the $D_B(\Delta P)$ and $D_E(\Delta P)$ measures, shown in Figure 8 using distribution plots for each direction separated by sequence length. The two directions in each plot have different patterns, but those patterns are reversed in the $D_B(\Delta P)$ and $D_E(\Delta P)$ variants: the $D_B(\Delta P_{RL})$ measure shows a high spike at the same location and with the same intensity as the $D_E(\Delta P_{LR})$ measure. This pattern is consistent across lengths, although the intensity of the spike and its location differs as length increases. The second, smaller spike represents meaningful sequences. For example, given the sequence [A – B – C – D], this spike represents the probability (A | BCD) and the probability (D | ABC). In both cases, associated sequences have high values while all other sequences have values near zero. The purpose of this figure is to show that the Divided class of measures differs from previous measures in that the relationship between left-to-right and right-to-left association also interacts with the variants of the measures, creating a mirrored distributional pattern. It remains the case, however, that directions of association need to be distinguished.

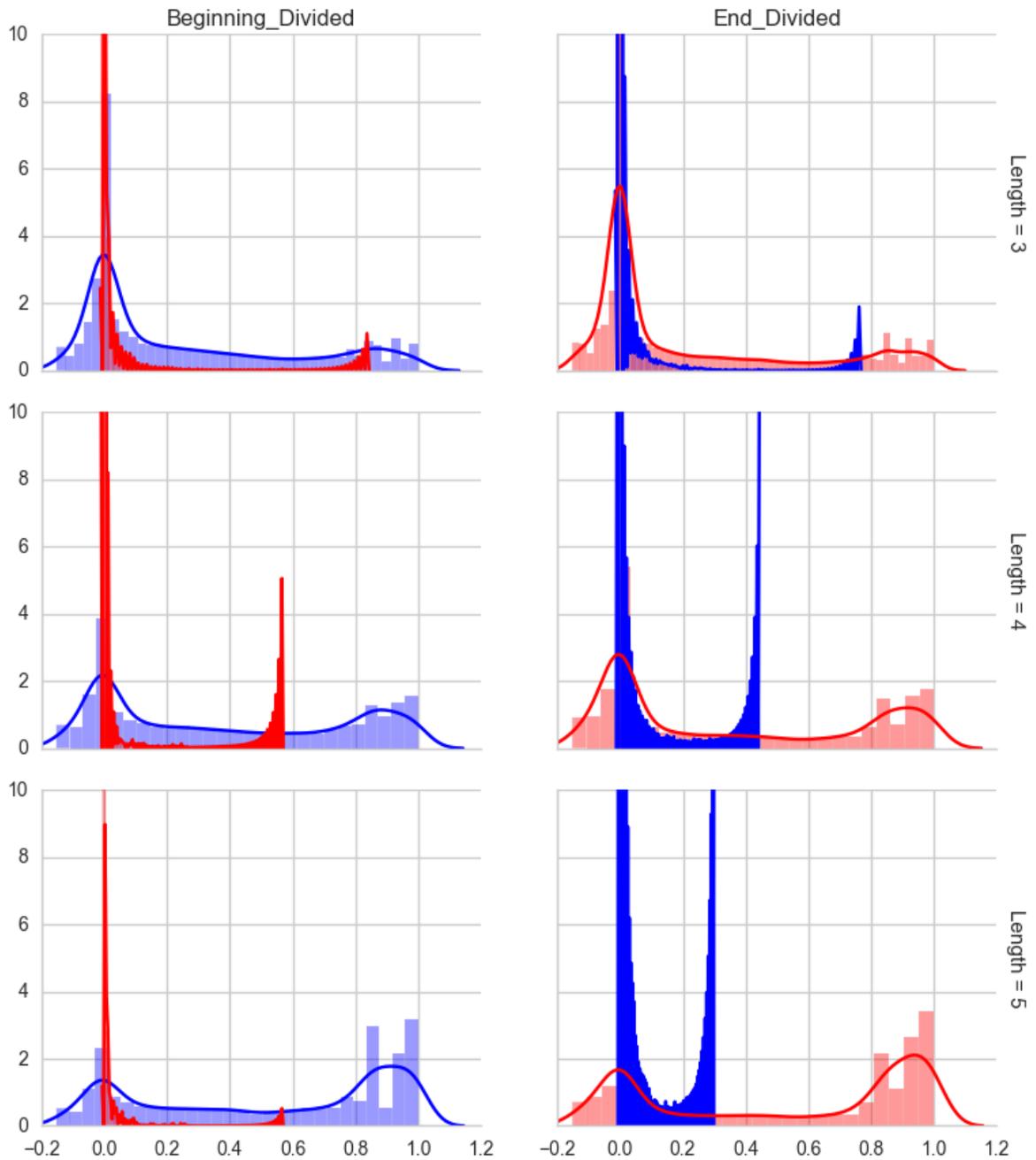

*Figure 8. Divided Measure Distribution Plots for English, LR (Blue) and RL (Red), By Length*

Now that we have reinforced previous findings of the importance of distinguishing between directions of association, we turn to the question of whether the multi-unit measures described in Section 2 overlap in their ranking of sequences: do these measures capture unique aspects of multi-unit association on a large-scale? The Spearman correlation between measures for English is shown in Table 15, with left-to-right association below the diagonal shaded in blue and right-to-left in italics above the diagonal shaded in green. Darker shades indicate higher correlation; the legend is shown below the table.

The only two measures that are very highly associated (above r = 0.75) are $\mu(\Delta P)$ and $\Sigma(\Delta P)$; this very high correlation holds in both directions. The other area of overlap is between $R_B(\Delta P)$ and $R_E(\Delta P)$, on the one hand, and $\mu(\Delta P)$ and $M(\Delta P)$, on the other hand (these measure are of course also correlated with $\Sigma(\Delta P)$). The correlations between these measures in both directions are around 0.50. This is because, as sequence length increases, the number of pairs contributing to the summed components of the Reduced class of measures also increases. The larger conclusion from these correlations, however, is that most of the measures produce different sequence rankings: each of the measures captures a particular pattern of multi-unit association and thus highlights aspects of association that may be missed by other measures.

*Table 15. Spearman Correlations Between Measures, LR (Below) and RL (Above)*

|   | $\Sigma(\Delta P)$ | $\mu(\Delta P)$ | $M(\Delta P)$ | $E(\Delta P)$ | $R_B(\Delta P)$ | $R_E(\Delta P)$ | $D_B(\Delta P)$ | $D_E(\Delta P)$ |
|---|---|---|---|---|---|---|---|---|
| $\Sigma(\Delta P)$ | -- | *.99* | *.69* | *.10* | *.55* | *.51* | *.10* | *.03* |
| $\mu(\Delta P)$ | .99 | -- | *.71* | *.10* | *.57* | *.53* | *.10* | *.03* |
| $M(\Delta P)$ | .72 | .73 | -- | *.15* | *.48* | *.52* | *.07* | *-.02* |
| $E(\Delta P)$ | .12 | .12 | .16 | -- | *.08* | *.15* | *.17* | *.09* |
| $R_B(\Delta P)$ | .57 | .59 | .56 | .12 | -- | *.02* | *.18* | *-.08* |
| $R_E(\Delta P)$ | .53 | .54 | .46 | .13 | .04 | -- | *.00* | *.17* |
| $D_B(\Delta P)$ | -.02 | -.01 | -.06 | .06 | .09 | -.10 | -- | *.41* |
| $D_E(\Delta P)$ | .10 | .10 | .07 | .13 | -.03 | .25 | .39 | -- |

*Table 15. Legend*

| **LR** .00 to .15 | .16 to .30 | .31 to .45 | .46 to .60 | .61 to .75 | .75 to .99 |
|---|---|---|---|---|---|
| **RL** .00 to .15 | .16 to .30 | .31 to .45 | .46 to .60 | .61 to .75 | .75 to .99 |

*5.2. Stability Across Languages and Representation Types*

How stable are these measures across languages and representations? This section looks at properties of the distribution of each measure as more direct evidence of cross-linguistic variation: each language has a different set of sequences, so we cannot compare ranks of sequences. Instead we compare their distributions. This is important for understanding the behavior of association measures. So far we have been examining lexical, syntactic, and mixed sequences together. Here we separate lexical and syntactic sequences. The question is whether measures of association are able to generalize across types of representation. To answer these questions, we compare each measure (left-to-right) on two conditions: first, using only lexical sequences; second, using only part-of-speech sequences.

Figure 9 shows each measure using three quantitative representations of its distribution, each plotted across eight languages. This allows us to compare changes in the distribution of the measures across both representation and language in a single setting. Kurtosis measures how well each measure matches the Gaussian distribution; negative values indicate the distribution is flatter and positive values indicate the distribution is more peaked than the Gaussian. Skewness measures how symmetrical the distribution of each measure is; high positive values indicate a right-tailed distribution and high negative values indicate a left-tailed distribution. Mean measures where the center of the distribution is located. We are not concerned here with how well each measure conforms to the Gaussian distribution along these dimensions. Rather, we are concerned with observing points of divergence across measures; the Gaussian distribution is used as a relative point of comparison to support the visualization of such divergences. To emphasize this, all measures are normalized to provide a scaled representation of their divergence.

For each measure, a flat line (across languages) indicates that a particular property of the distribution is consistent. The distance between the red and blue points indicates whether a particular property of the distribution is consistent across lexical and syntactic sequences. While there are many language-specific and measure-specific observations that could be made using Figure 9, for example that French syntactic sequences are an outlier for their mean value, we focus instead on consistency across languages and representations in order to identify areas in which results from smaller studies may be insufficient.

Kurtosis, the degree to which a distribution is peaked, is consistent across representation types and languages for most measures. The two outliers are Italian lexical sequences for $E(\Delta P)$ and Spanish syntactic sequences for $D_E(\Delta P)$, both of which have significantly higher peaks. This means that these two categories of sequences are more heavily centered around their mean values. In the first case, this means that Italian lexical sequences are less likely to have associated end-points (because the mean here is zero). In the second case, this means that Spanish syntactic sequences are less likely to have their core components predict the right end-point. Beyond these two exceptions, however, we see that the measures remain consistent across languages and representation types in their kurtosis.

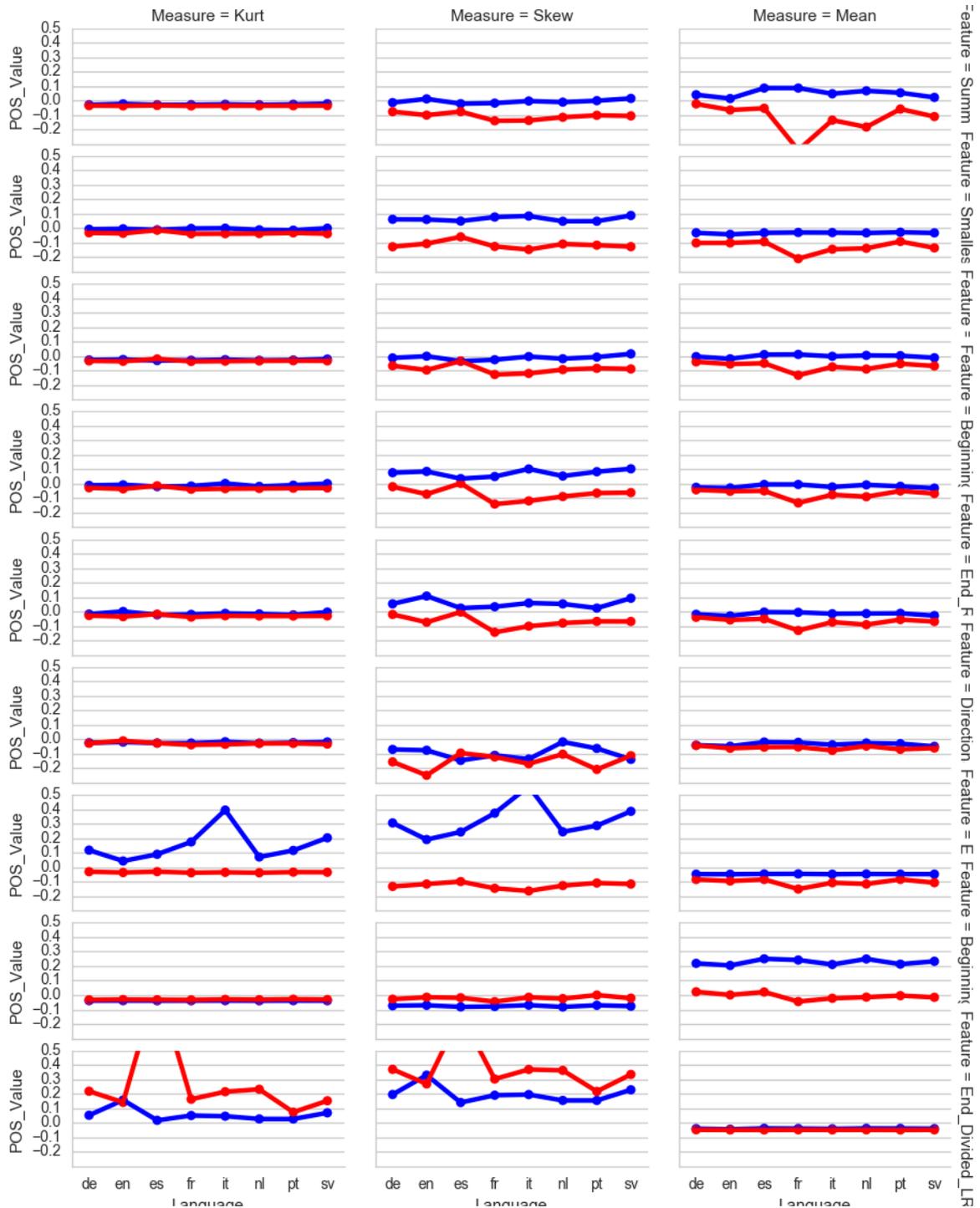

*Figure 9. Kurtosis, Skew, and Mean Across Languages for LR: Lexical (Blue) and Syntactic (Red)*

Skewness, the degree to which a distribution is right-tailed or left-tailed, is also relatively consistent across languages, although here we see a slight separation between lexical and syntactic sequences: in all cases except the Divided class, lexical sequences have a more right-tailed distribution than syntactic sequences. The two exceptions are the same as before: Italian lexical sequences for *E(ΔP)* and Spanish syntactic sequences for *D$_E$(ΔP)*. The Italian lexical sequences for the *E(ΔP)* are not as much an outlier here: lexical sequences across languages are significantly more right-tailed than syntactic sequences for this measure. This means that there are fewer sequences with high values for *E(ΔP)* for lexical sequences, an outcome that makes intuitive sense because many phrases like *blue paper pizza* have lexical end-points that are unlikely to co-occur but syntactic end-points that are likely to co-occur (i.e., ADJECTIVE – NOUN).

Mean, the center value of the distribution, is more consistent across languages and conditions than skew, with most plots being flat with very close lexical and syntactic sequences. There are three exceptions to this: First, the mean value of syntactic sequences is lower for *μ(ΔP)*, indicating that syntactic sequences are generally less associated. Second, French syntactic sequences are an outlier in many of the measures, showing a generally lower mean than syntactic sequences for other languages. Third, *D$_B$(ΔP)*, while consistent across languages, has a significantly lower mean for syntactic sequences in general.

This sort of analysis is important because we want to generalize these measures across languages and representations, but this requires that the measures are relatively consistent in their behavior. Many studies do not cover multiple languages so that each of the exceptions noted above would be viewed as a measure-specific variation on smaller datasets. As shown in the external resources accompanying this paper, the influence of frequency weighting and of using the unadjusted conditional probability as the base measure are also consistent across languages. This shows that these measures generalize well across conditions.

## 6. Conclusions

The motivation for this paper has been to generalize association measures across varying sequence lengths and levels of representation. The problem is that generalizing across different lengths creates segmentation problems and generalizing across type of representation creates very large numbers of sequences. Both the qualitative analysis (Section 4) and the large-scale quantitative analysis (Section 5) suggest that the measures developed here capture different aspects of multi-unit association. This implies that we are unlikely to find a single measure that captures all facets of multi-unit association. The current approach produces a vector of association values for each sequence, one or more of which can reveal meaningful or otherwise interesting collocations. For example, in Section 4 we used the $M(\Delta P)$ and $C_C(\Delta P)$ measures to filter results from other measures, thus combining multiple measures in a simple way. This is an expanded version of earlier suggestions of using tuples of association, frequency, dispersion, and entropy (Gries 2012). The studies in this paper strongly suggest that a vector-based representation is important once we leave behind pairs of lexical items for sequences of varying lengths and levels of representation.

     A vector-based approach complicates the use of association measures because we now have 16 measures producing 16 distinct sequence rankings. In order to make sense of these measures, we take up the idea of filtering in Table 16 by presenting a list of top LR and RL sequences produced by combining the measures into a single direction-specific feature ranking. In order to filter sequences, first, we have 'constraint' measures that must be satisfied: Sequences that have weak links are removed from the ranking; this is defined as an $M(\Delta P)$ that falls below 0.01. Sequences that have shifting directions of association are also removed from the ranking; this is defined as an $C_C(\Delta P)$ greater than 1. Second, we have 'ranking' measures: for the remaining sequences, we represent each one using the its highest direction-specific measure. For example, if a sequence has an $E$ of 0.04, an $R_B$ of 0.004, and an $D_B$ of 0.005, then it is represented using $E$, the measure which has the maximum value across all individual measures representing that sequence. This results in the sequences shown in Table 16.

*Table 16. Top Sequences Filtered By a Vector of Association Values*

| LR Sequences | RL Sequences |
|---|---|
| the european union | able to VERB DETERMINER |
| passenger name record | credit default swaps |
| globalisation adjustment fund | uncompetitive coal mines |
| data protection supervisor | gross domestic product |
| internal market scoreboard | reversed qualified majority |
| general motors belgium | herbal medicinal products |
| agree with DETERMINER rapporteur | sovereign wealth funds |
| genetically modified soya | nordic passport union |
| illegally staying third-country | disputes arising from |
| mont blanc tunnel | i would like to |

The rankings in Table 16 illustrate a simple way to use a vector of association values: if each measure reveals a different aspect of association, we can use them all together to provide a more comprehensive picture of associated sequences. This table also supports two simple observations: First, although syntactic units within sequences greatly increases the number of sequences total (c.f., Figure 4), syntactic sequences are very common and thus tend to have lower association values than lexical items. Second, in both directions the sequences fall into multiple linguistic categories: complex noun phrases (e.g., *the european union*), complex verb phrases (e.g., *able to VERB DETERMINER*), and partially-fixed argument structures (e.g., *disputes arising from*, *i would like to*). This range of sequence types shows the robustness of a vector-based approach to association.

To put vectors of association measures into a wider context, how do they compare with word embeddings (e.g., Erhan, et al. 2010; Pennington, et al., 2014)? First, the items defined are sequences rather than single units. Second, and more importantly, word embeddings are relations between units or sequences and their context (i.e., skip-grams along dimensions of common lexical items for each unit) while vectors of association values are relations between units in a sequence regardless of context. Thus, it is plausible to conceive of a two-staged approach in which (first) a vector of association values is used to identify those sequences which are interesting or meaningful, and then (second) word embeddings are used to measure the similarity of the contexts in which these sequences occur. In short, association values indicate whether and how a sequence is meaningful while word embeddings indicate whether and how individual units are meaningful; there are many relationships between these two corpus-based representations that deserve further exploration.

**Appendix 1: Comparing the Unweighted and the Frequency Weighted ΔP**

This section examines the difference between the raw ΔP and the frequency-weighted ΔP: how does this change the overall distribution of sequences? We are interested in this comparison because the combination of association and frequency has the potential to resolve theoretical conflicts between the relative importance of these two types of measures (e.g., Bybee 2006; Gries 2012). On the other hand, by merging both measures together (through multiplication), frequency weighting represents sequence A that has an association score of 0.9 and a co-occurrence frequency of 50 in exactly the same way that it represents sequence B that has an association score of 0.045 and a co-occurrence frequency of 1000. We have seen in the main paper itself that frequency weighting has a different qualitative effect for each measure; but that analysis considered only the top ten lexical sequences for each measure in a single language. Here we look at all sequences across eight languages.

Our first approach to the comparison is to look at the agreement in the ranking of sequences with and without frequency weighting using Spearman correlations. High correlations mean that the conditions rank sequences in a similar way but low correlations mean that there is a difference that needs to be investigated further. This is shown in Figure 1 with a heat map of correlation strength, with darker cells indicating higher agreement. Rows indicate measures, so that each row shows the correlation between weighted and unweighted values for a single measure across languages. Our first question is whether languages are consistent: does frequency weighting have the same influence for a given measure across all languages, or does it have a different influence across languages? In Figure 1 we see that measures dominate, with little variation across languages. For example, the Σ(ΔP) and μ(ΔP) measures show little agreement between rankings: frequency weighting changes the order of sequences. However, the low correlations are consistent so that we do not have to worry about language-specific effects. Some cross-linguistic variation remains: the Σ(ΔP$_{LR}$) measure has lower agreement for Spanish (0.658) and higher agreement for French (0.727). But this is a much smaller difference than variations across measures. This and later figures visualize a large number of values which would be too cumbersome to display individually. However, the full results are available as external resources.

[*Figure A1. Correlation Between Weighted and Unweighted Measures Across Languages*]

From Figure 1 we see that nine measures have high agreement between sequence rankings: $R_B(\Delta P)$ and $R_E(\Delta P)$ in both directions, $C_C(\Delta P)$, $E(\Delta P)$, $D_B(\Delta P_{RL})$, and $D_E(\Delta P_{LR})$. We do not need to explore these measures more closely. Two more measures have medium-high agreement between rankings and will also not be explored these more closely: $M(\Delta P)$ in both directions. In Figure 2 we compare the $\Sigma(\Delta P)$ and $\mu(\Delta P)$ measures (the Divided class of measures is an interesting case and we pick up this question in a later section). The figure shows the unweighted ΔP (in blue) and the frequency-weighted ΔP (in red) using a normalized distribution plot; normalization is required here because the two conditions have very different ranges. In this figure the measures are columns with each row showing the same measure in a different language. The question we want to answer with this figure is *how* frequency weighting influences the distribution of sequences for those measures in which it does have an influence.

*[Figure A2. Normalized Distribution of Raw (Blue) and Frequency Weighted (Red) Measures]*

The basic generalization is that the unweighted ΔP measures have a distribution much more peaked around a few values (i.e., most association values cluster around 0) while the frequency-weighted measures are more evenly distributed (i.e., there are more values that are very positive or very negative). The summed weighted values have a particularly large range. The distributions across languages are similar, which again indicates an absence of language-specific effects. Here we also see that the LR and RL distributions are quite similar, so that there are no direction-specific effects influencing the distribution that need to be explored further.

The purpose of the analysis displayed in Figures 1 and 2 is to identify where frequency weighting has a strong influence and to determine whether this influence in consistent across languages. The conclusion is that it does have a consistent influence on some measures, specifically $\mu(\Delta P)$. The initial conclusions from the English examples discussed in the main paper indicate that the unweighted measure favors sequences that may be rare but which always occur together in the dataset (i.e., named entities such as *Porto Alegre*). The weighted measure, however, favors sequences that are both associated and contain individual units that are highly frequent (i.e., idiomatic phrases such as *in order to*). These sequences likely have lower association, in the sense that "in" occurs in many other collocations, but are promoted by their sheer frequency.

This raises two considerations: First, which measure should we use? The answer here depends on the task: if we want to find named entities, then the unweighted measures seem to perform better; if we want to find grammaticalized sequences, however, the frequency weighted measures seem more appropriate. Second, what is the cognitive status of frequency weighted association? Are there other methods of combining association and frequency that correspond better to a cognitive process that language learners use to grammaticalize structure from observed usage? While this is a matter for future work, one approach is to employ both sorts of measures for the task of learning grammatical structures and evaluate which produces the more accurate representations. For example, if frequency weighted association consistently reveals grammaticalized structures more clearly than raw association, this would provide one piece of evidence that frequency and association, combined in this way, have a certain cognitive reality. This is a question for future work, however, and the purpose here is to identify where and how robustly these conditions differ in order to identify where such future work should focus.

## Appendix 2: Comparing the ΔP with Conditional Probability

The pairwise ΔP that forms the core of each of the multi-unit measures subtracts the conditional probability of one unit occurring without the other from the conditional probability of both units occurring together. The next task is to examine the influence that this adjustment has: how would the behavior of these measures change if we simply used the conditional probability itself as the core measure? We start by looking at the similarity in sequence ranks between these two conditions, using a heat map of Spearman correlations across languages in Figure A3. The point of this visualization is to reveal those contexts in which the conditions differ and which thus merit further examination. We again see relative consistency across languages, with the variation occurring across measures. In this case, the only measures showing low agreement are the *M(ΔP)* and the *E(ΔP)*, in both directions.

[*Figure A3. Correlation Between ΔP and Conditional Probability Across Languages]*

The *ΔP* controls for the presence of the outcome without the cue. Another way of looking at this adjustment is that it controls for the baseline probability of the second unit occurring after any generic unit in the corpus. This baseline creates negative values for cases in which the current pair co-occurs less frequently than the baseline. For example, given the sequence *give me,* a negative value for the *ΔP* would indicate that *me* is less likely to follow *give* than any random unit in the corpus. It is not surprising that this adjustment has a significant influence on the *M(ΔP)*, then, because this adjustment highlights the presence of weak links. The low correlation between conditions for this measure shows that the *ΔP* is actually doing what it is meant to do: reveal cases in which observed association is accidental.

The other measure in which the conditions differ is both directions of *E(ΔP)*, which is meant to find sequences that have fixed end-points but variable internal units. The particularly low correlation here is because the *ΔP* takes on negative values when the end-points are not frequently observed together. In these cases, the probability of the outcome without the cue is much higher than the probability of the outcome with the cue. Again, this is a scenario in which the *ΔP* excels at measuring the particular property that is highlighted by the *E(ΔP)* measure.

In the end, then, the *ΔP* and the conditional probability are quite similar except in cases where the *ΔP* excels in not over-estimating the attraction between units. This pattern is stable across languages, as before, which gives us confidence that the *ΔP* actually does provide an improved core measure rather than just exploiting a property particular to the English data on which it has previously been evaluated.

*Figure A1. Correlation Between Weighted and Unweighted Measures Across Languages*

| | de | en | es | fr | it | nl | pt | sv |
|---|---|---|---|---|---|---|---|---|
| $\Sigma(\Delta P_{LR})$ | | | | | | | | |
| $M(\Delta P_{LR})$ | | | | | | | | |
| $\Sigma(\Delta P_{RL})$ | | | | | | | | |
| $M(\Delta P_{RL})$ | | | | | | | | |
| $\mu(\Delta P_{LR})$ | | | | | | | | |
| $\mu(\Delta P_{RL})$ | | | | | | | | |
| $R_B(\Delta P_{LR})$ | | | | | | | | |
| $R_B(\Delta P_{RL})$ | | | | | | | | |
| $R_E(\Delta P_{LR})$ | | | | | | | | |
| $R_E(\Delta P_{RL})$ | | | | | | | | |
| $C_S(\Delta P)$ | | | | | | | | |
| $C_C(\Delta P)$ | | | | | | | | |
| $E(\Delta P_{LR})$ | | | | | | | | |
| $E(\Delta P_{RL})$ | | | | | | | | |
| $D_B(\Delta P_{LR})$ | | | | | | | | |
| $D_B(\Delta P_{RL})$ | | | | | | | | |
| $D_E(\Delta P_{LR})$ | | | | | | | | |
| $D_E(\Delta P_{RL})$ | | | | | | | | |

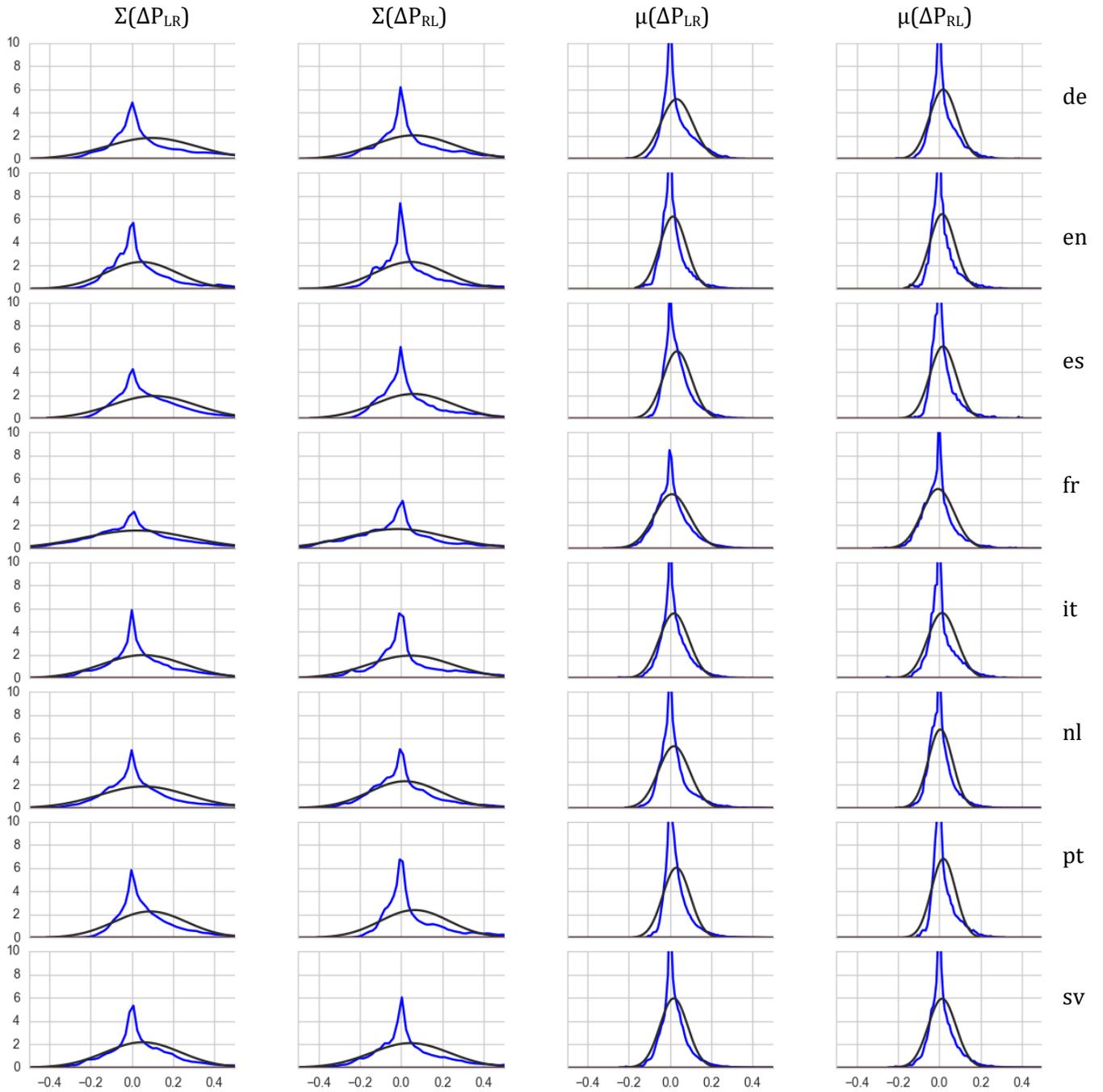

*Figure A2. Normalized Distribution of Raw (Blue) and Frequency Weighted (Red) Measures*

*Figure A3. Correlation Between ΔP and Conditional Probability Across Languages*

|  | de | en | es | fr | it | nl | pt | sv |
|---|---|---|---|---|---|---|---|---|
| Σ(ΔP$_{LR}$) | | | | | | | | |
| M(ΔP$_{LR}$) | | | | | | | | |
| Σ(ΔP$_{RL}$) | | | | | | | | |
| M(ΔP$_{RL}$) | | | | | | | | |
| μ(ΔP$_{LR}$) | | | | | | | | |
| μ(ΔP$_{RL}$) | | | | | | | | |
| R$_B$(ΔP$_{LR}$) | | | | | | | | |
| R$_B$(ΔP$_{RL}$) | | | | | | | | |
| R$_E$(ΔP$_{LR}$) | | | | | | | | |
| R$_E$(ΔP$_{RL}$) | | | | | | | | |
| D$_S$ | | | | | | | | |
| D$_C$ | | | | | | | | |
| E(ΔP$_{LR}$) | | | | | | | | |
| E(ΔP$_{RL}$) | | | | | | | | |
| D$_B$(ΔP$_{LR}$) | | | | | | | | |
| D$_B$(ΔP$_{RL}$) | | | | | | | | |
| D$_E$(ΔP$_{LR}$) | | | | | | | | |
| D$_E$(ΔP$_{RL}$) | | | | | | | | |